\documentclass[11pt,a4paper]{article}
\usepackage[hyperref]{emnlp2018}
\usepackage{times,graphicx,dingbat}
\usepackage{latexsym,booktabs,multicol,multirow}
\usepackage{amsmath}
\usepackage{pifont}
\newcommand{\xmark}{\ding{55}}%
\usepackage{booktabs}
\usepackage[font={small}]{caption}

\aclfinalcopy 
\def\aclpaperid{1373} 

\usepackage{url,colortbl}

\title{Parameter sharing between dependency parsers for related languages}

\author{Miryam de Lhoneux$^{1}$\thanks{\hspace{0.7em}Work carried out during a stay at the University of Copenhagen.}
~~~Johannes Bjerva$^{2}$~~~Isabelle Augenstein$^{2}$~~~Anders S\o gaard$^{2}$\\[1mm] 
\begin{tabular}{cp{10mm}c}
    $^{1}$Department of Linguistics and Philology && $^{2}$ Department of Computer Science \\ 
    Uppsala University && University of Copenhagen \\ 
    Uppsala, Sweden && Copenhagen, Denmark \\
\end{tabular}
}

\date{}

\begin{document}
\maketitle
\begin{abstract}
Previous work has suggested that parameter sharing between transition-based neural dependency parsers for related languages can lead to better performance, but there is no consensus on what parameters to share. We present an evaluation of 27 different parameter sharing strategies across 10 languages, representing five pairs of related languages, each pair from a different language family. We find that sharing transition classifier parameters always helps, whereas the usefulness of sharing word and/or character LSTM parameters varies. Based on this result, we propose an architecture where the transition classifier is shared, and the sharing of word and character parameters is controlled by a parameter that can be tuned on validation data. This model is linguistically motivated and obtains significant improvements over a mono-lingually trained baseline. We also find that sharing transition classifier parameters helps when training a parser on unrelated language pairs, but we find that, in the case of unrelated languages, sharing too many parameters does not help.
\end{abstract}

\section{Introduction}
The idea of sharing parameters between parsers of related languages goes back to early work in cross-lingual adaptation \cite{Zeman:Resnik:08}, and the idea has  recently received a lot of interest in the context of neural dependency parsers \cite{Duong:ea:15,Ammar:ea:16,Susanto:Lu:17}. Modern neural dependency parsers, however, use different sets of parameters for representation and scoring, and it is not clear what parameters it is best to share. 

The Universal Dependencies (UD) project \citep{nivre16universal}, which is seeking to harmonize the annotation of dependency treebanks across languages, has seen a steady increase in languages that have a treebank in a common standard. Many of these languages are low resource and have small UD treebanks. It seems interesting to find out ways to leverage the wealth of information contained in these treebanks, especially for low resource languages.

In this paper, we evaluate 27 different parameter sharing strategies. We focus on a particular transition-based neural dependency parser \cite{delhoneux17raw,delhoneux17arc}, which performs close to the state of the art. This parser has three sets of parameters: i) the parameters of a character-based one-layer, bidirectional LSTM; ii) the parameters of a word-based two-layer, bidirectional LSTM; iii) and the parameters of a multi-layered perceptron (MLP) with a single hidden layer. The two first sets are for learning to represent configurations; the third for selecting the next transition. We consider all combinations of sharing these sets of parameters; and in addition, we consider two ways of sharing each set of parameters, namely with and without a prefixed language embedding. The latter enables partial, soft sharing. In sum, we consider all $3^3$ combinations of no sharing, hard sharing and soft sharing of the three sets of parameters. 
We evaluate the 27 multilingual parsers on 10 languages from the UD project, representing five pairs of related languages, each pair from a different language family. We repeat the experiment with the same set of languages, but using pairs of unrelated languages.

\paragraph{Contributions} This paper is, to the best of our knowledge, the first to evaluate different parameter sharing strategies for exploiting synergies between neural dependency parsers of related languages. 
We evaluate the different strategies on 10 languages, representing five different language families. We find that sharing (MLP) transition classifier parameters always helps, whereas the usefulness of sharing LSTM parameters depends on the language pair. This reflects the intuition that the transition classifier learns hierarchical structures that are likely to transfer across languages, based on parser configurations that abstract away from several linguistic differences. The similarity of the input to character- and word-level LSTMs, on the other hand, will vary depending on the phonological and morphosyntactic similarity of the languages in question. Motivated by this observation, we propose an architecture with hard-wired transition classifier parameter sharing, but in which sharing of LSTM parameters is tuned. The novel architecture significantly outperforms our monolingual baseline on our set of 10 languages.
We additionally investigate parameter sharing of unrelated languages.

\begin{table}\begin{center}\begin{tabular}{lrlll}
\toprule
{\bf Lang}&{\bf Tokens}&{\bf Family}&{\bf Word order} \\
\midrule ar & 208,932 & Semitic & VSO \\
he& 161,685 & Semitic & SVO \\
et& 60,393 & Finnic & SVO \\
fi& 67,258 & Finnic & SVO \\
hr& 109,965 & Slavic & SVO \\
ru& 90,170 & Slavic   & SVO \\
it& 113,825 & Romance & SVO \\
es& 154,844 & Romance & SVO \\
nl& 75,796 & Germanic & No dom. order \\
no& 76,622 & Germanic & SVO \\
\bottomrule
\end{tabular}\end{center}
\caption{\label{data}Dataset characteristics}\end{table}

\section{The Uppsala dependency parser}
The Uppsala parser \cite{delhoneux17raw,delhoneux17arc} consists of three sets of parameters; the parameters of the character-based LSTM, those of the word-based LSTM, and the parameters of the MLP that predicts transitions. The character-based LSTM produces representations for the word-based LSTM, which produces representations for the MLP. The Uppsala parser is a transition-based parser \citep{kiperwasser16}, adapted to the Universal Dependencies (UD) scheme,\footnote{\url{http://universaldependencies.org/}} and using the arc-hybrid transition system from \citet{kuhlmann11} extended with a \textsc{Swap} transition and a static-dynamic oracle, as described in \citet{delhoneux17arc}. The \textsc{Swap} transition is used to generate non-projective dependency trees \citep{nivre09acl}. 

For an input sentence of length $n$ with words $w_1,\dots,w_n$, the parser creates a sequence of vectors $x_{1:n}$, where the vector $x_i$ representing $w_i$ is the concatenation of a word embedding and the final state of the character-based LSTM after processing the characters of $w_i$. The character vector $ch(w_i)$ is obtained by running a (bi-directional) LSTM over the characters $ch_j$ ($1 \leq j \leq m$) of $w_i$. 
Each input element is represented by the word-level, bi-directional LSTM, as a vector $v_i = \textsc{BiLstm}(x_{1:n},i)$. For each configuration, the feature extractor concatenates the LSTM representations of core elements from the stack and buffer. Both the embeddings and the LSTMs are trained together with the model.

A configuration $c$ is represented by a feature function $\phi(\cdot)$ over a subset of its elements. For each configuration, transitions are scored by a classifier, in this case an MLP, and $\phi(\cdot)$ is a concatenation of \mbox{BiLSTM} vectors on top of the stack and the beginning of the buffer. The MLP scores transitions together with the arc labels for transitions that involve adding an arc. In practice, we use two interpolated MLPs, one which only scores the transitions, and one which scores transitions together with the arc label. For simplicity, we refer to that interpolated MLP as the MLP.

\begin{table*}\begin{center}
\begin{tabular}{l|ccc|cc|cc|cc|cc|cc|c}
\toprule
{\bf Model}&{\bf C}&{\bf W}&{\bf S}&ar&he&es&it&et&fi&nl&no&hr&ru&{\sc Av}\\
\midrule
{\sc Mono}&&&& 76.3 & 80.2 & 83.7 & 83.3 & 70.4 & 70.8 & 77.3 & 80.8 & 76.8 & 82.3 & 78.2  \\
\midrule
\multicolumn{4}{c}{\sc Language-best}&76.6	& 80.6	& 84.4	& 84.8	& 72.8	& 72.9	& 79.6	& 82.1	& 78.0	& 82.9	& 79.5\\
\midrule
{\sc Best}& \xmark & \checkmark & ID          & 76.3 & 80.3 & 84.2 & 84.5 & 72.1 & 72.5 & 78.8 & 81.4 & 77.6 & 82.8 & 79.1    \\
\midrule
{\sc Char}& \checkmark & \xmark     & \xmark     & 76.4 & 80.3 & 84.3 & 84.0 & 72.3 & 71.0 & 78.3 & 81.3 & 77.0 & 82.3 & 78.7    \\
{\sc Word}& \xmark     & \checkmark & \xmark     & 76.3 & 79.9 & 83.9 & 84.4 & 72.4 & 71.3 & 77.4 & 80.7 & 76.9 & 82.5 & 78.6    \\
{\sc State}& \xmark     & \xmark     & \checkmark & 76.6 & 80.3 & 84.0 & 83.7 & 71.5 & 72.9 & 78.3 & 81.5 & 77.4 & 82.8 & 78.9    \\
\multicolumn{4}{c}{\vdots}\\
{\sc All}& \checkmark & \checkmark & \checkmark & 76.2 & 80.1 & 84.0 & 84.2 & 72.1 & 71.4 & 78.7 & 81.1 & 77.0 & 82.5 & 78.7    \\
{\sc Soft}& ID         & ID         & ID         & 76.3 & 79.9 & 84.1 & 84.4 & 72.1 & 71.3 & 79.6 & 81.4 & 77.1 & 82.5 & 78.9    \\
\bottomrule
\end{tabular}\caption{\label{sharing}Performance on development data (LAS; in \%) across select sharing strategies. {\sc Mono} is our single-task baseline; {\sc Language-best} is using the best sharing strategy for each language (as evaluated on development data); {\sc Best} is the overall best sharing strategy, across languages; {\sc Char} shares only the character-based LSTM parameters; {\sc Word} shares only the word-based LSTM parameters; {\sc All} shares all parameters. \checkmark~refers to hard sharing, ID refers to soft sharing, using an embedding of the language ID and \xmark~refers to not sharing.}
\end{center}\end{table*}

\section{Parameter sharing}

Since our parser has three basic sets of model parameters, we consider sharing all combinations of those three sets.
We also introduce two ways of sharing, namely, with or without the addition of a vector representing the language. This language embedding enables the model, in theory, to learn what to share between the two languages in question. Since for all three model parameter sets, we now have three options -- not sharing, sharing, or sharing in the context of a language embedding -- we are left with $3^3=27$ parameter sharing strategies; see Table~\ref{sharing}.

In the setting where we do not share (\xmark) word parameters ({\bf W}), we construct a different word lookup table and a different word-level BiLSTM for each language. In the setting where we do hard parameter sharing (\checkmark) of word parameters, we only construct one lookup table and one word BiLSTM for the languages involved. In the setting where we do soft sharing (ID) of word parameters, we share those parameters, and in addition, concatenate a language embedding $l_i$ representing the language of word $w_i$ to the vector of the word $w_i$ at the input of the word BiLSTM: $ x_i = e(w_i) \circ ch(w_i) \circ l_i$. Similarly for character parameters ({\bf C}), we construct a different character BiLSTM and one character lookup for each language (\xmark), create those for all languages and share them (\checkmark) or share them and concatenate a (randomly initialized) language embedding $l_i$ representing the language of word $w_i$ at the input of the character BiLSTM (ID): $ch_j = e(ch_j) \circ l_i$. At the level of configuration or parser states ({\bf S}), we either construct a different MLP for each language (\xmark), share the MLP (\checkmark) or share it and concatenate a language embedding $l_i$ representing the language of word $w_i$ to the vector representing the configuration, at the input of the MLP (ID): $ c = \phi(\cdot) \circ l_i$.

\section{Experiments}

\paragraph{Language pairs}
We use 10 languages in our experiments, representing five language pairs from different language families. Our two {\sc Semitic} languages are Arabic and Hebrew. These two languages differ in that Arabic tends to favour VSO word order whereas Hebrew tends to use SVO, but are similar in their rich transfixing morphology. Our two {\sc Finno-Ugric} languages are Estonian and Finnish. These two languages differ in that Estonian no longer has vowel harmony, but share a rich agglutinative morphology. Our two {\sc Slavic} languages are Croatian and Russian. These two languages differ in that Croatian uses gender in plural nouns, but otherwise share their rich inflectional morphology. Our two {\sc Romance} languages are Italian and Spanish. These two languages differ in that Italian uses a possessive adjective with a definite article, but share a fairly strict SVO order. Finally, our two {\sc Germanic} languages are Dutch and Norwegian. These two languages differ in morphological complexity, but share word ordering features to some extent.

\paragraph{Datasets} For all 10 languages, we use treebanks from the Universal Dependencies project. The dataset characteristics are listed in Table~\ref{data}. To keep the results comparable across language pairs, we down-sample the training set to the size of the smallest of our languages, Hebrew: we randomly sample 5000 sentences for each training set. 
Note that while this setting makes the experiment somewhat artificial and will probably overestimate the benefits that can be obtained from sharing parameters when using larger treebanks, we find it interesting to see how much low resource languages can benefit from parameter sharing, as explained in the introduction.

\paragraph{Baselines and systems} This is an evaluation paper, and our results are intended to explore a space of sharing strategies to find better ways of sharing parameters between dependency parsers of related languages. Our baseline is the Uppsala parser trained monolingually. Our systems are parsers trained bilingually by language pair where we share subsets of parameters between the languages in the pair, and we report on what sharing strategies seem superior across the 10 languages that we consider. 
\paragraph{Implementation details} A flexible implementation of parameter strategies for the Uppsala parser was implemented in Dynet.\footnote{\url{https://github.com/clab/dynet}} We make the code publicly available.\footnote{\url{https://github.com/coastalcph/uuparser}}

\section{Results and discussion}

Our results on development sets are presented in Table~\ref{sharing}. We use labeled attachment score (LAS) as our metric for evaluating parsers. Table~\ref{sharing} presents numbers for a select subset of the 27 sharing strategies. The other results can be found in the supplementary material. Our main observations are: {\bf (i)} that, generally, and as observed in previous work, {\em multi-task learning helps}: all different sharing strategies are on average better than the monolingual baselines, with minor (0.16 LAS points) to major (0.86 LAS points) average improvements; and {\bf (ii)} that sharing the MLP seems to be overall a better strategy than not sharing it: the 10 best strategies share the MLP. Whereas the usefulness of sharing the MLP seems to be quite robust across language pairs, the usefulness of sharing word and character parameters seems more dependent on the language pairs. This reflects the linguistic intuition that character- and word-level LSTMs are highly sensitive to phonological and morphosyntactic differences such as word order, whereas the MLP learns to predict less idiosyncratic, hierarchical relations from relatively abstract representations of parser configurations. \\ 
\indent Based on this result, we propose a model ({\sc Ours}) where {\em the MLP is shared and the sharing of word and character parameters is controlled by a parameter that can be set on validation data}. 
Results are given in Table~\ref{test}. We obtain a 0.6 LAS improvement on average and our proposed model is significantly better than the monolingual baseline with $p<0.01$. Significance testing is performed using a randomization test, with the script from the CoNLL 2017 Shared Task.\footnote{\url{https://github.com/udapi/udapi-python/blob/master/udapi/block/eval/conll17.py}}

\begin{table}
    \centering
    \begin{tabular}{l|ll|ccc}
    \toprule
    & {\bf W}&{\bf C}& \sc{Ours} & {\sc Mono}& $\delta$ \\
        \midrule
        ar  & \xmark     & \xmark     & 77.2 & 77.1 & 0.1    \\
        es  & ID         & \checkmark & 84.3 & 83.8 & 0.5    \\
        et  & \xmark     & ID         & 71.4 & 70.5 & 0.8    \\
        fi  & \xmark     & \xmark     & 71.6 & 71.6 & 0.1    \\
        he  & \checkmark & \xmark     & 80.0 & 79.8 & 0.3    \\
        hr  & \checkmark & \xmark     & 77.9 & 78.0 & -0.1   \\
        it  & ID         & \checkmark & 85.0 & 84.0 & 1.0    \\
        nl  & ID         & \checkmark & 75.5 & 74.1 & 1.4    \\
        no  & \xmark     & ID         & 81.1 & 80.1 & 1.0    \\
        ru  & \checkmark & \xmark     & 83.5 & 82.7 & 0.8    \\
        \midrule
        av. &            &            & 78.8 & 78.2 & 0.6   \\
        \bottomrule
    \end{tabular}
    \caption{LAS on the test sets of the best of 9 sharing strategies and the monolingual baseline. $\delta$ is the difference between \sc{Ours} and \sc{Mono}.}
    \label{test}
\end{table}

\section{Unrelated languages}
We repeated the same set of experiments with unrelated language pairs. We hypothesise that parameter sharing between unrelated language pairs will be less useful in general than with related language pairs. However, it can still be useful, it has been shown previously that unrelated languages can benefit from being trained jointly. For example, \citet{lynn14cross} have shown that Indonesian was surprisingly particularly useful for Irish.\\
\indent The results are presented in Table~\ref{unrelated}. The table only presents part of the results, the rest can be found in the supplementary material.
\indent As expected, there is much less to be gained from sharing parameters between unrelated pairs. However, it is possible to improve the monolingual baseline by sharing some of the parameters. In general, sharing the MLP is still a helpful thing to do. It is most helpful to share the MLP and optionally one of the two other sets of parameters. Results are close to the monolingual baseline when everything is shared. Sharing word and character parameters but not the MLP hurts accuracy compared to the monolingual baseline.

\begin{table*}
    \begin{center}
    \begin{tabular}{l|ccc|cc|cc|cc|cc|cc|c}
        \multicolumn{1}{l|}{{\bf Model}} & {\bf C}    & {\bf W}    & {\bf S}    & he   & no   & fi   & hr   & ru   & es   & it   & et   & nl   & ar   & \textsc{Av} \\ \hline
        \multicolumn{1}{l|}{{\sc Mono}}  &            &            &            & 80.2 & 80.8 & 70.8 & 76.8 & 82.3 & 83.7 & 83.3 & 70.4 & 77.3 & 76.3 & 78.2    \\ \hline
        \multicolumn{4}{c}{\sc Language-best}                                   & 80.5 & 81.5 & 71.9 & 77.6 & 82.9 & 84.0 & 84.3 & 72.5 & 78.7 & 76.5 & 78.9    \\ \hline
        \multicolumn{1}{l|}{{\sc Best}}  & \xmark     & \xmark     & \checkmark & 80.3 & 81.5 & 71.9 & 77.6 & 82.7 & 84.0 & 83.8 & 72.5 & 78.7 & 76.3 & 78.9    \\ 
        \multicolumn{1}{l|}{{\sc Worst}}  & ID         & ID         & \xmark     & 79.8 & 80.6 & 69.2 & 76.7 & 81.4 & 83.8 & 83.2 & 69.4 & 76.6 & 76.0 & 77.7    \\ \hline
        \multicolumn{1}{l|}{{\sc Char}}  & \checkmark & \xmark     & \xmark     & 80.1 & 80.9 & 71.4 & 76.8 & 82.9 & 83.9 & 84.3 & 70.9 & 78.0 & 76.5 & 78.6    \\
        \multicolumn{1}{l|}{{\sc Word}}  & \xmark     & \checkmark & \xmark     & 79.6 & 80.9 & 71.9 & 76.9 & 82.2 & 83.7 & 83.8 & 70.9 & 77.0 & 76.4 & 78.3    \\
        \multicolumn{1}{l|}{{\sc All}}   & \checkmark & \checkmark & \checkmark & 80.5 & 80.9 & 69.8 & 76.6 & 82.3 & 83.7 & 84.0 & 70.6 & 77.4 & 76.2 & 78.2    \\
        \multicolumn{1}{l|}{{\sc Soft}}  & ID         & ID         & ID         & 79.8 & 80.5 & 70.1 & 76.6 & 82.1 & 83.9 & 83.8 & 70.6 & 77.2 & 76.3 & 78.1   
    \end{tabular}
    \caption{\label{unrelated}Performance on development data (LAS; in \%) across select sharing strategies for unrelated languages. {\sc Mono} is our single-task baseline; {\sc Language-best} is using the best sharing strategy for each language (as evaluated on development data); {\sc Best} and {\sc Worst} are the overall best and worst sharing strategy across languages; {\sc Char} shares only the character-based LSTM parameters; {\sc Word} shares only the word-based LSTM parameters; {\sc All} shares all parameters. \checkmark~refers to hard sharing, ID refers to soft sharing, using an embedding of the language ID and \xmark~refers to not sharing.}
    \end{center}
\end{table*}

\section{Related work}

Previous work has shown that sharing parameters between dependency parsers for related languages can lead to improvements \cite{Duong:ea:15,Ammar:ea:16,Susanto:Lu:17}. \citet{smith2018st} recently found that sharing parameters using the same parser as in this paper (soft sharing of word parameters, hard sharing of the rest) improves parsing accuracy when training on related languages, and is especially useful in the low resource case. Similar effects have been observed in machine translation \cite{Dong:ea:15,Johnson:ea:17}, for example. Most studies have only explored a small number of parameter sharing strategies, however.  
\citet{vilares16one} evaluate parsing with hard parameter sharing for 100 language pairs with a statistical parser. \citet{naseem12selective} proposed to selectively share subsets of a parser across languages in the context of a probabilistic parser.\\
\indent Options we do not explore here are learning the architecture jointly with optimizing the task objective \cite{Misra2016,ruder2017sluice}, or learning an architecture search model that predicts an architecture based on the properties of datasets, typically with reinforcement learning~\cite{Zoph:Le:17,Wong:Gesmundo:18,Liang:ea:18}. 
We also do not explore the option of sharing selectively based on more fine-grained typological information about languages, which related work has indicated could be useful \cite{N18-1083}. Rather, we stick to sharing between languages of the same language families.

The strategies explored here do not exhaust the space of possible parameter sharing strategies. For example, we completely ignore soft sharing based on mean-constrained regularisation \cite{Duong:ea:15}. 

\section{Conclusions}

We present evaluations of 27 parameter sharing strategies for the Uppsala parser across 10 languages, representing five language pairs from five different language families. We repeated the experiment with pairs of unrelated languages. We made several observations: (a) Generally, multi-task learning helps. (b) Sharing the MLP parameters always helps. It helps to share MLP parameters when training a parser on a pair of related languages, and it also helps if the languages are unrelated. (c) Sharing word and character parameters is differently helpful depending on the language. (d) Sharing too many parameters does not help, when the languages are unrelated.\\
\indent In future work, we plan to investigate what happens when training on more than 2 languages. Here, we focused on a setting with rather small amounts of balanced data. It would be interesting to experiment with using datasets that are not balanced with respect to size. 
Finally, we have restricted our experiments to a specific architecture, using fixed hyperparameters including word and character embedding dimensions. It would be interesting to experiment with different parsing architectures as well as varying those hyperparameters. 

\section*{Acknowledgments}

We acknowledge the computational resources provided by CSC in Helsinki and Sigma2 in Oslo through NeIC-NLPL (www.nlpl.eu).
The second author is partially funded by an AdeptMind scholarship. The last author was funded by an ERC Starting Grant.

\bibliography{emnlp2018}

\begin{thebibliography}{22}
\expandafter\ifx\csname natexlab\endcsname\relax\def\natexlab#1{#1}\fi

\bibitem[{Ammar et~al.(2016)Ammar, Mulcaire, Ballesteros, Dyer, and
  Smith}]{Ammar:ea:16}
Waleed Ammar, George Mulcaire, Miguel Ballesteros, Chris Dyer, and Noah~A.
  Smith. 2016.
\newblock More languages, one parser.
\newblock In \emph{TACL}.

\bibitem[{Bjerva and Augenstein(2018)}]{N18-1083}
Johannes Bjerva and Isabelle Augenstein. 2018.
\newblock {From Phonology to Syntax: Unsupervised Linguistic Typology at
  Different Levels with Language Embeddings}.
\newblock In \emph{Proceedings of the 2018 Conference of the North American
  Chapter of the Association for Computational Linguistics: Human Language
  Technologies, Volume 1 (Long Papers)}, pages 907--916. Association for
  Computational Linguistics.

\bibitem[{Dong et~al.(2015)Dong, Wu, He, Yu, and Wang}]{Dong:ea:15}
Daxiang Dong, Hua Wu, Wei He, Dianhai Yu, and Haifeng Wang. 2015.
\newblock Multi-task learning for multiple language translation.
\newblock In \emph{ACL}.

\bibitem[{Duong et~al.(2015)Duong, Cohn, Bird, and Cook}]{Duong:ea:15}
Long Duong, Trevor Cohn, Steven Bird, and Paul Cook. 2015.
\newblock {Low Resource Dependency Parsing: Cross-lingual Parameter Sharing in
  a Neural Network Parser}.
\newblock In \emph{Proceedings of ACL}.

\bibitem[{Johnson et~al.(2017)Johnson, Schuster, Le, Krikun, Wu, Chen, Thorat,
  Viégas, Wattenberg, Corrado, Hughes, and Dean}]{Johnson:ea:17}
Melvin Johnson, Mike Schuster, Quoc~V. Le, Maxim Krikun, Yonghui Wu, Zhifeng
  Chen, Nikhil Thorat, Fernanda Viégas, Martin Wattenberg, Greg Corrado,
  Macduff Hughes, and Jeffrey Dean. 2017.
\newblock Google's neural machine translation system.
\newblock In \emph{TACL}.

\bibitem[{Kiperwasser and Goldberg(2016)}]{kiperwasser16}
Eliyahu Kiperwasser and Yoav Goldberg. 2016.
\newblock {Simple and Accurate Dependency Parsing Using Bidirectional {LSTM}
  Feature Representations}.
\newblock \emph{TACL}, 4:313--327.

\bibitem[{Kuhlmann et~al.(2011)Kuhlmann, G\'{o}mez-Rodr\'{i}guez, and
  Satta}]{kuhlmann11}
Marco Kuhlmann, Carlos G\'{o}mez-Rodr\'{i}guez, and Giorgio Satta. 2011.
\newblock {Dynamic Programming Algorithms for Transition-Based Dependency
  Parsers}.
\newblock In \emph{Proceedings of ACL}, pages 673--682, Portland, Oregon, USA.

\bibitem[{de~Lhoneux et~al.(2017{\natexlab{a}})de~Lhoneux, Shao, Basirat,
  Kiperwasser, Stymne, Goldberg, and Nivre}]{delhoneux17raw}
Miryam de~Lhoneux, Yan Shao, Ali Basirat, Eliyahu Kiperwasser, Sara Stymne,
  Yoav Goldberg, and Joakim Nivre. 2017{\natexlab{a}}.
\newblock From raw text to universal dependencies - look, no tags!
\newblock In \emph{{Proceedings of the CoNLL 2017 Shared Task: Multilingual
  Parsing from Raw Text to Universal Dependencies}}, pages 207--217, Vancouver,
  Canada.

\bibitem[{de~Lhoneux et~al.(2017{\natexlab{b}})de~Lhoneux, Stymne, and
  Nivre}]{delhoneux17arc}
Miryam de~Lhoneux, Sara Stymne, and Joakim Nivre. 2017{\natexlab{b}}.
\newblock {Arc-Hybrid Non-Projective Dependency Parsing with a Static-Dynamic
  Oracle}.
\newblock In \emph{Proceedings of the 15th International Conference on Parsing
  Technologies}, pages 99--104, Pisa, Italy.

\bibitem[{Liang et~al.(2018)Liang, Meyerson, and Miikkulainen}]{Liang:ea:18}
Jason Liang, Elliot Meyerson, and Risto Miikkulainen. 2018.
\newblock {Evolutionary Architecture Search For Deep Multitask Networks}.
\newblock In \emph{GECCO}.

\bibitem[{Lynn et~al.(2014)Lynn, Foster, Dras, and Tounsi}]{lynn14cross}
Teresa Lynn, Jennifer Foster, Mark Dras, and Lamia Tounsi. 2014.
\newblock Cross-lingual transfer parsing for low-resourced languages: An irish
  case study.
\newblock In \emph{Proceedings of the First Celtic Language Technology
  Workshop}, pages 41--49.

\bibitem[{Misra et~al.(2016)Misra, Shrivastava, Gupta, and Hebert}]{Misra2016}
Ishan Misra, Abhinav Shrivastava, Abhinav Gupta, and Martial Hebert. 2016.
\newblock {Cross-Stitch Networks for Multi-Task Learning}.
\newblock In \emph{Proceedings of CVPR}.

\bibitem[{Naseem et~al.(2012)Naseem, Barzilay, and
  Globerson}]{naseem12selective}
Tahira Naseem, Regina Barzilay, and Amir Globerson. 2012.
\newblock Selective sharing for multilingual dependency parsing.
\newblock In \emph{Proceedings of the 50th Annual Meeting of the Association
  for Computational Linguistics (Volume 1: Long Papers)}, pages 629--637.
  Association for Computational Linguistics.

\bibitem[{Nivre(2009)}]{nivre09acl}
Joakim Nivre. 2009.
\newblock {Non-Projective Dependency Parsing in Expected Linear Time}.
\newblock In \emph{Proceedings of ACL}, pages 351--359, Suntec, Singapore.

\bibitem[{Nivre et~al.(2016)Nivre, de~Marneffe, Ginter, Goldberg, Hajic,
  Manning, McDonald, Petrov, Pyysalo, Silveira et~al.}]{nivre16universal}
Joakim Nivre, Marie-Catherine de~Marneffe, Filip Ginter, Yoav Goldberg, Jan
  Hajic, Christopher~D Manning, Ryan McDonald, Slav Petrov, Sampo Pyysalo,
  Natalia Silveira, et~al. 2016.
\newblock Universal dependencies v1: A multilingual treebank collection.
\newblock In \emph{Proceedings of the 10th International Conference on Language
  Resources and Evaluation (LREC 2016)}.

\bibitem[{Ruder et~al.(2017)Ruder, Bingel, Augenstein, and
  S{\o}gaard}]{ruder2017sluice}
Sebastian Ruder, Joachim Bingel, Isabelle Augenstein, and Anders S{\o}gaard.
  2017.
\newblock {Sluice networks: Learning what to share between loosely related
  tasks}.
\newblock In \emph{CoRR, abs/1705.08142}.

\bibitem[{Smith et~al.(2018)Smith, Bohnet, de~Lhoneux, Nivre, Shao, and
  Stymne}]{smith2018st}
Aaron Smith, Bernd Bohnet, Miryam de~Lhoneux, Joakim Nivre, Yan Shao, and Sara
  Stymne. 2018.
\newblock {82 Treebanks, 34 Models: Universal Dependency Parsing with
  Multi-Treebank Models}.
\newblock In \emph{{Proceedings of the CoNLL 2018 Shared Task: Multilingual
  Parsing from Raw Text to Universal Dependencies}}.

\bibitem[{Susanto and Lu(2017)}]{Susanto:Lu:17}
Raymond~Hendy Susanto and Wei Lu. 2017.
\newblock Neural architectures for multilingual semantic parsing.
\newblock In \emph{ACL}.

\bibitem[{Vilares et~al.(2016)Vilares, G{\'o}mez-Rodr{\'\i}guez, and
  Alonso}]{vilares16one}
David Vilares, Carlos G{\'o}mez-Rodr{\'\i}guez, and Miguel~A Alonso. 2016.
\newblock One model, two languages: training bilingual parsers with harmonized
  treebanks.
\newblock In \emph{Proceedings of the 54th Annual Meeting of the Association
  for Computational Linguistics (Volume 2: Short Papers)}, volume~2, pages
  425--431.

\bibitem[{Wong and Gesmundo(2018)}]{Wong:Gesmundo:18}
Catherine Wong and Andrea Gesmundo. 2018.
\newblock {Transfer Learning to Learn with Multitask Neural Model Search}.
\newblock In \emph{ICPR}.

\bibitem[{Zeman and Resnik(2008)}]{Zeman:Resnik:08}
Daniel Zeman and Philip Resnik. 2008.
\newblock {Cross-Language Parser Adaptation between Related Languages}.
\newblock In \emph{IJCNLP}.

\bibitem[{Zoph and Le(2017)}]{Zoph:Le:17}
Barret Zoph and Quoc Le. 2017.
\newblock {Neural Architecture Search with Reinforcement Learning}.
\newblock In \emph{ICPR}.

\end{thebibliography}
\bibliographystyle{acl_natbib_nourl}

\appendix

\begin{table*}[ht]
    \centering
        \begin{tabular}{ccc|cc|cc|cc|cc|cc|c}
            \textbf{C}         & \textbf{W}         & \textbf{S} &  ar   & he   & es   & it   & et   & fi   & nl   & no   & hr   & ru   & average \\
        \hline
        \sc{Mono}   & & &  76.3 & 80.2 & 83.7 & 83.3 & 70.4 & 70.8 & 77.3 & 80.8 & 76.8 & 82.3 & 78.2  \\
        \hline
        \xmark     & \checkmark & ID         & 76.3 & 80.3 & 84.2 & 84.5 & 72.1 & 72.5 & 78.8 & 81.4 & 77.6 & 82.8 & 79.1    \\
        \xmark     & \checkmark & \checkmark & 76.4 & 80.4 & 84.1 & 84.4 & 71.9 & 72.0 & 78.7 & 81.5 & 78.0 & 82.8 & 79.0    \\
        \xmark     & ID         & ID         & 76.2 & 80.1 & 84.2 & 84.8 & 71.8 & 72.4 & 78.6 & 81.6 & 77.4 & 82.9 & 79.0    \\
        \xmark     & ID         & \checkmark & 76.4 & 80.4 & 84.2 & 84.3 & 72.2 & 72.3 & 78.3 & 81.6 & 77.4 & 82.8 & 79.0    \\
        \xmark     & \xmark     & ID         & 76.5 & 80.6 & 84.1 & 84.3 & 71.8 & 72.0 & 78.5 & 81.5 & 77.7 & 82.9 & 79.0    \\
        \checkmark & ID         & \checkmark & 76.2 & 79.8 & 84.4 & 84.7 & 72.4 & 71.5 & 79.2 & 81.6 & 76.9 & 82.8 & 78.9    \\
        ID         & \xmark     & \checkmark & 76.3 & 80.2 & 84.0 & 84.4 & 72.8 & 71.7 & 78.6 & 82.1 & 77.2 & 82.3 & 78.9    \\
        \xmark     & \xmark     & \checkmark & 76.6 & 80.3 & 84.0 & 83.7 & 71.5 & 72.9 & 78.3 & 81.5 & 77.4 & 82.8 & 78.9    \\
        ID         & ID         & ID         & 76.3 & 79.9 & 84.1 & 84.4 & 72.1 & 71.3 & 79.6 & 81.4 & 77.1 & 82.5 & 78.9    \\
        \checkmark & ID         & ID         & 76.2 & 80.1 & 84.3 & 84.5 & 72.5 & 71.8 & 78.7 & 81.1 & 76.7 & 82.6 & 78.8    \\
        ID         & \xmark     & \xmark     & 76.6 & 80.3 & 84.1 & 84.3 & 71.8 & 71.7 & 78.3 & 81.5 & 77.2 & 82.5 & 78.8    \\
        ID         & \xmark     & ID         & 76.2 & 79.9 & 84.1 & 84.3 & 71.9 & 72.0 & 78.5 & 81.7 & 77.2 & 82.3 & 78.8    \\
        \xmark     & \xmark     & \xmark     & 76.6 & 80.3 & 84.0 & 84.2 & 71.2 & 72.1 & 78.0 & 81.6 & 77.3 & 82.6 & 78.8    \\
        \checkmark & \xmark     & \checkmark & 76.1 & 80.1 & 84.0 & 84.1 & 72.1 & 72.0 & 78.2 & 81.8 & 76.8 & 82.6 & 78.8    \\
        \checkmark & \xmark     & ID         & 76.5 & 80.0 & 84.0 & 84.4 & 71.9 & 71.7 & 78.5 & 81.7 & 76.7 & 82.3 & 78.8    \\
        \checkmark & \checkmark & ID         & 76.6 & 80.1 & 84.0 & 84.6 & 72.1 & 71.0 & 78.3 & 81.0 & 76.9 & 82.8 & 78.7    \\
        ID         & ID         & \checkmark & 76.1 & 80.1 & 84.1 & 84.6 & 72.1 & 71.2 & 78.0 & 81.4 & 77.0 & 82.7 & 78.7    \\
        \checkmark & \xmark     & \xmark     & 76.4 & 80.3 & 84.3 & 84.0 & 72.3 & 71.0 & 78.3 & 81.3 & 77.0 & 82.3 & 78.7    \\
        \checkmark & \checkmark & \checkmark & 76.2 & 80.1 & 84.0 & 84.2 & 72.1 & 71.4 & 78.7 & 81.1 & 77.0 & 82.5 & 78.7    \\
        \xmark     & ID         & \xmark     & 76.6 & 80.1 & 84.1 & 84.3 & 71.7 & 71.6 & 77.6 & 81.0 & 77.0 & 82.6 & 78.7    \\
        ID         & \checkmark & \checkmark & 76.2 & 79.9 & 83.8 & 84.5 & 72.4 & 70.3 & 78.1 & 81.0 & 77.2 & 82.6 & 78.6    \\
        \xmark     & \checkmark & \xmark     & 76.3 & 79.9 & 83.9 & 84.4 & 72.4 & 71.3 & 77.4 & 80.7 & 76.9 & 82.5 & 78.6    \\
        ID         & \checkmark & ID         & 76.0 & 80.0 & 83.8 & 84.3 & 71.7 & 70.9 & 78.3 & 81.0 & 76.9 & 82.5 & 78.5    \\
        \checkmark & \checkmark & \xmark     & 76.1 & 79.7 & 83.8 & 84.5 & 71.9 & 70.4 & 77.8 & 81.1 & 76.5 & 82.3 & 78.4    \\
        \checkmark & ID         & \xmark     & 76.0 & 79.3 & 84.1 & 84.4 & 71.5 & 71.3 & 77.7 & 80.6 & 76.7 & 82.5 & 78.4    \\
        ID         & \checkmark & \xmark     & 76.1 & 79.9 & 83.9 & 84.4 & 71.1 & 70.6 & 77.8 & 80.9 & 77.0 & 82.0 & 78.4    \\
        ID         & ID         & \xmark     & 75.9 & 79.5 & 84.1 & 84.4 & 72.1 & 70.5 & 77.4 & 80.6 & 77.0 & 82.2 & 78.4  
    \end{tabular}
    \caption{\label{sharingfull} Performance (LAS; in \%) across select sharing strategies ranked by average performance. {\sc Mono} is our single-task baseline; \textbf{W} refers to sharing the word parameters, \textbf{C} refers to sharing the character parameters, \textbf{S} refers to sharing the MLP parameters. \checkmark~refers to hard sharing, ID refers to soft sharing, using an embedding of the language ID and \xmark~refers to not sharing. }
\end{table*}

\begin{table*}[ht]
    \centering
    \begin{tabular}{ccc|cc|cc|cc|cc|cc|c}
        \textbf{C} & \textbf{W} & \textbf{S} & he   & no   & fi   & hr   & ru   & es   & it   & et   & nl   & ar   & average \\
        \hline
        \sc{Mono} & &  & 80.2 & 80.8 & 70.8 & 76.8 & 82.3 & 83.7 & 83.3 & 70.4 & 77.3 & 76.3 & 78.2  \\
        \hline
        \xmark     & \xmark     & \checkmark & 80.3 & 81.5 & 71.9 & 77.6 & 82.7 & 84.0 & 83.8 & 72.5 & 78.7 & 76.4 & 78.9    \\
        \xmark     & \xmark     & \xmark     & 80.3 & 81.7 & 71.9 & 77.7 & 82.5 & 84.0 & 84.2 & 71.8 & 78.5 & 76.5 & 78.9    \\
        \xmark     & ID         & ID         & 80.3 & 81.1 & 72.1 & 77.7 & 82.7 & 84.2 & 84.2 & 72.4 & 77.9 & 76.0 & 78.9    \\
        \xmark     & \xmark     & ID         & 79.7 & 81.5 & 72.2 & 77.1 & 82.7 & 83.8 & 84.0 & 72.3 & 78.6 & 76.5 & 78.8    \\
        ID         & \xmark     & ID         & 80.3 & 81.6 & 71.8 & 77.5 & 82.6 & 84.1 & 83.8 & 71.8 & 78.3 & 76.2 & 78.8    \\
        ID         & \xmark     & \xmark     & 80.0 & 81.6 & 71.5 & 77.2 & 82.7 & 83.9 & 84.0 & 71.1 & 79.0 & 76.4 & 78.7    \\
        \checkmark & \xmark     & ID         & 80.3 & 81.5 & 71.5 & 77.5 & 82.7 & 83.7 & 83.9 & 71.6 & 77.9 & 76.7 & 78.7    \\
        ID         & \xmark     & \checkmark & 80.4 & 81.4 & 71.6 & 77.3 & 82.6 & 83.9 & 84.1 & 71.9 & 77.7 & 76.4 & 78.7    \\
        \xmark     & \checkmark & ID         & 80.5 & 81.2 & 72.2 & 77.0 & 82.5 & 84.0 & 83.8 & 71.5 & 78.3 & 76.1 & 78.7    \\
        \xmark     & ID         & \checkmark & 80.6 & 81.1 & 71.9 & 77.1 & 82.7 & 84.2 & 84.0 & 71.9 & 77.1 & 76.2 & 78.7    \\
        \checkmark & \xmark     & \checkmark & 80.3 & 81.1 & 71.3 & 77.0 & 82.6 & 84.0 & 84.2 & 71.8 & 77.8 & 76.3 & 78.6    \\
        \xmark     & \checkmark & \checkmark & 80.6 & 81.1 & 71.6 & 76.8 & 82.5 & 83.7 & 83.9 & 71.4 & 78.1 & 76.3 & 78.6    \\
        \checkmark & \xmark     & \xmark     & 80.1 & 80.9 & 71.4 & 76.8 & 82.9 & 83.9 & 84.3 & 70.9 & 78.0 & 76.5 & 78.6    \\
        \xmark     & ID         & \xmark     & 80.3 & 81.3 & 71.2 & 77.4 & 81.9 & 84.2 & 83.9 & 70.7 & 77.6 & 75.8 & 78.4    \\
        \xmark     & \checkmark & \xmark     & 79.6 & 80.9 & 71.9 & 76.9 & 82.2 & 83.7 & 83.8 & 70.9 & 77.0 & 76.4 & 78.3    \\
        ID         & \checkmark & ID         & 80.3 & 81.0 & 70.5 & 76.5 & 82.3 & 83.7 & 83.6 & 71.4 & 77.8 & 76.1 & 78.3    \\
        ID         & \checkmark & \checkmark & 80.1 & 80.8 & 70.4 & 77.0 & 82.2 & 83.8 & 83.8 & 71.0 & 77.6 & 76.2 & 78.3    \\
        \checkmark & \checkmark & \checkmark & 80.5 & 80.9 & 69.8 & 76.6 & 82.3 & 83.7 & 84.0 & 70.6 & 77.4 & 76.2 & 78.2    \\
        \checkmark & ID         & ID         & 80.3 & 80.7 & 70.2 & 76.1 & 82.1 & 83.8 & 83.8 & 70.8 & 77.6 & 76.2 & 78.2    \\
        \checkmark & \checkmark & ID         & 79.8 & 80.9 & 71.0 & 76.2 & 82.1 & 83.7 & 83.6 & 70.9 & 77.3 & 76.0 & 78.2    \\
        ID         & ID         & \checkmark & 80.0 & 80.8 & 69.8 & 76.2 & 82.2 & 83.8 & 84.3 & 70.7 & 77.2 & 76.2 & 78.1    \\
        ID         & ID         & ID         & 79.8 & 80.5 & 70.1 & 76.6 & 82.1 & 83.9 & 83.8 & 70.6 & 77.2 & 76.3 & 78.1    \\
        \checkmark & ID         & \checkmark & 80.3 & 81.1 & 70.2 & 76.1 & 82.2 & 84.1 & 83.7 & 70.3 & 76.9 & 76.0 & 78.1    \\
        \checkmark & \checkmark & \xmark     & 80.4 & 80.3 & 70.1 & 76.6 & 82.0 & 83.7 & 83.2 & 69.3 & 76.7 & 76.2 & 77.8    \\
        \checkmark & ID         & \xmark     & 79.6 & 80.6 & 69.4 & 76.7 & 81.7 & 83.8 & 83.4 & 69.2 & 77.6 & 76.2 & 77.8    \\
        ID         & \checkmark & \xmark     & 79.6 & 80.1 & 69.9 & 76.6 & 81.8 & 83.5 & 82.9 & 69.2 & 77.6 & 76.3 & 77.7    \\
        ID         & ID         & \xmark     & 79.8 & 80.6 & 69.2 & 76.7 & 81.4 & 83.8 & 83.2 & 69.4 & 76.6 & 76.0 & 77.7   

    \end{tabular}
    \caption{\label{sharingfull} Unrelated language pairs. Performance (LAS; in \%) across select sharing strategies ranked by average performance. {\sc Mono} is our single-task baseline; \textbf{W} refers to sharing the word parameters, \textbf{C} refers to sharing the character parameters, \textbf{S} refers to sharing the MLP parameters. \checkmark~refers to hard sharing, ID refers to soft sharing, using an embedding of the language ID and \xmark~refers to not sharing. }
\end{table*}

\end{document}